\documentclass[11pt,letterpaper]{article}
\usepackage{eamt15}
\usepackage{times}
\usepackage{latexsym}
\usepackage{url}
\usepackage{booktabs}
\usepackage{color}
\usepackage{etex}
\usepackage{times}
\usepackage{latexsym}
\usepackage{amsmath}
\usepackage{amssymb}
\usepackage{graphicx}
\usepackage{subcaption}
\usepackage{multirow}
\usepackage{adjustbox}
\usepackage{titlesec}
\usepackage{url}
\makeatletter
\g@addto@macro{\UrlBreaks}{\UrlOrds}
\makeatother
\usepackage{enumitem}
\usepackage{float}
\usepackage{verbatim}
\usepackage{booktabs}
\usepackage{tikz}
\usepackage{fancyref}

\newcommand{\mc}[3]{\multicolumn{#1}{#2}{#3}}

\title{Resolving Out-of-Vocabulary Words with Bilingual Embeddings in Machine Translation}

\author{Pranava Swaroop Madhyastha ~~~~\t {\bf Cristina Espa\~na-Bonet}\\
 TALP Research Center \\  Univesitat Polit\`ecnica de Catalunya\\ Jordi Girona, 1-3, 08034 Barcelona, Spain \\
\{pranava,cristinae\}@cs.upc.edu }

\date{}

\begin{document}

\maketitle

\begin{abstract}
Out-of-vocabulary words account for a large proportion of errors in machine
translation systems, especially when the system is 
used on a different domain than the one where it was trained. 
In order to alleviate the problem, we propose to use a log-bilinear softmax-based model for
vocabulary expansion, such that given an out-of-vocabulary source word, the model
generates a probabilistic list of possible translations in the target language.
Our model uses only word embeddings trained on significantly large unlabelled monolingual 
corpora and trains over a fairly small, word-to-word bilingual
dictionary.
% ~\footnote{System, models and compressed embeddings would be made
% available.}.
We input this probabilistic list into a standard phrase-based statistical machine
translation system and
obtain consistent improvements in translation quality on the English--Spanish
language pair. Especially, we get an improvement of 3.9 BLEU points when
tested over an out-of-domain testset.
\end{abstract}

\section{Introduction} 
\label{s:intro}

%\red{(verbose, friendly version, a lot to cut!)}

Data-driven machine translation systems are able to translate words that have been seen 
in the training corpora, however translating unseen words is still a major challenge 
for even the best performing systems. 
% As the amount of parallel data is finite (and sometimes scarce) named entities, spe-
% cialised words in a new domain or simply infre-
% quent terms will be absent in the system, and the
% lack of any information about these items can po-
% tentially result in a bad translation.
In general, the amount of parallel data is finite (and sometimes scarce) which
results in word types like named entities, domain specific content words, or 
infrequent terms to be absent in the training parallel corpora.
This lack of information can potentially result in 
incomplete or erroneous translations.

This area has been actively studied in the field of machine translation
(MT)~\cite{habash:2008,daume:2011,marton:2009,rapp:1999,dou:2012,irvine:2013}. 
%In most cases, it has been seen~\cite{rasooli:2013,durrani:2014} that the
%system when trained with a large
%parallel corpus and if restricted to the same domain, 
%most of the unseen words will probably be named
%entities. Previous research suggests that significantly large number of 
%named entities can be handled by using simple
%pre-or-post-processing, like transliteration methods~\cite{hermjakob2008name,al:2002}.
%However, a change in domain results in a significant increase in the number
%of unseen words. These unseen words might include a significant proportion of 
%regular domain specific content words. 
Lexicon based resources have been used for resolving unseen content words by
exploiting a combination of monolingual and bilingual
resources~\cite{rapp:1999,Callison-BurchEtal:2006,SalujaEtal:2014,zhaoEtal:2015}.
In this context, distributed word representations, or word embeddings (WE), have been
recently applied to resolve unseen word related
problems~\cite{mikolovEtal:2013,zouEtal:2013}. In general,  word
representations capture rich linguistic relationships. Several
works~\cite{bilbowa,wuEtal:2014} try to use WE to improve MT systems. 
However, very few approaches use them directly to resolve the out-of-vocabulary (OOV)
problem for MT systems.

Our work is inspired by the recent advances in applications of word embeddings
to the task of vocabulary expansion in the context of statistical machine translation (SMT). 
In this work, we introduce a principled method to obtain a probabilistic distribution 
of words in the target language for a given source word.
We do this by using WEs in both languages and
learning a log-bilinear softmax model that is trained using a relatively small
bilingual lexicon (the seed lexicon) to obtain a probabilistic distribution of words.
%%\red{(please, rephrase!)}
%b) i
Then, we integrate the generated distribution of target words for every
unseen source word into a standard SMT system.

%In this work we specifically work on improving the
%translation of OOVs. We train monolingual embeddings on significantly large
%corpora. We use a sufficiently large parallel corpus to train a MT system. We
%also use two different settings for learning language models - one built on the
%parallel corpora and the other using an expanded corpus and learn a good
%language model. Further, we evaluate on both settings, i.e., with a standard
%setting and over an out of domain dataset. We obtain consistent improvement 

The rest of the paper is organised as follows.
In the next section we briefly describe some previous related work. 
Section~\ref{s:bwe} presents the log-bilinear softmax model, and its integration into an 
SMT system and the SMT experiments sre analysed in Section~\ref{s:exp}. 
Finally, in Section 5, we draw our conclusions and sketch some future work.

\section{Background and Motivation}
There are several strands of related research that try to alleviate the effect
of unseen words in translation. Previous research suggests that a significantly large number of 
named entities can be handled by using simple
pre/post-processing, like transliteration methods~\cite{hermjakob2008name,al:2002}.
However, a change in domain results in a significant increase in the number
of unseen words. These unseen words might include a significant proportion of 
regular domain-specific content words. 

Our focus in this paper is to resolve unseen content words by using continuous word
embeddings on both the languages and a small seed lexicon to map the embeddings. 
To this extent, our work is similar to~\newcite{ishiwatariinstant}
where the authors map distributional representations using a linear regression method
similar to~\newcite{mikolovEtal:2013} and insert a new feature based on cosine
similarity metric into the MT system. In our work, we use a principled method
to obtain a probabilistic conditional distribution of words directly and these 
probabilities allow us to expand the translation model for the new words. 

There are other related works~\cite{rapp:1999,daume:2011,durrani:2014} that have explored
 approaches based
on extracting lexicons using corpus based methods to resolve out of training
vocabulary problems. There is also a rich body of recent literature that focuses on obtaining
bilingual word embeddings using either sentence aligned corpora or document
aligned corpora~\cite{Klementiev:2012,bilbowa,Kocisky:2014}. Our approach is
significantly different as we obtain embeddings separately on monolingual corpora and
then use supervision in the form of a small sparse bilingual dictionary.

\section{Mapping Continuous Word Represen- tations using a Bilinear Model}
\label{s:bwe}

\paragraph{Definitions.}
Let $\mathcal{E}$ and $\mathcal{F}$ be the vocabularies of the two languages, source and target, and
let $e \in \mathcal{E}$ and $f \in \mathcal{F}$ be the words in the languages respectively. We are given 
with a relatively small set of source word to target word $e \to f$ dictionary. We also
assume that we have access to some kind of distributed word embeddings in both 
languages. Let $\phi(.)\to \mathbb{R}^n$ denote the $n$-dimensional distributed
representation of the words, and let us assume we have both source ($\phi_{s}$)
    and target ($\phi_{t}$) embeddings. The task we are interested in is
to learn a model for the conditional probability distribution $\Pr(e{\mid}f)$. That is,
given a word in a source language, say English $(e)$, we want to get a conditional probability 
distribution of all the words in a foreign language $(f)$. 
% We use out of vocabulary or OOV alternatively with unseen words. 

\paragraph{Log-Bilinear Softmax Model.}
We look at this task as a bilinear prediction task as proposed
by~\newcite{madhyastha14} and extend it for the bilingual setting. The proposed
model makes use of word embeddings on both
languages with no additional features. The basic function is formulated as
log-bilinear softmax model and takes the following form: 
\begin{equation}
\Pr(f|e;W) =
\frac{\exp\{\phi_{s}(e)^{\top}W\phi_{t}(f)\}}{\sum_{f' \in
\mathcal{F}}\exp\{\phi_{s}(e)^{\top}W\phi_{t}(f')\}}
\end{equation}
Essentially, our problem reduces to: a) first getting the corresponding word embeddings of 
the vocabularies on both the languages on a significantly large 
monolingual corpus and b) estimating $W$ given a relatively small dictionary.
That is, to learn $W$ we use the source word to 
target word dictionary as training supervision.

We learn $W$ by minimizing the negative log-likelihood of the dictionary using a 
nuclear norm regularized objective as: $L(W) = - \sum_{s,t}\log(\Pr(t|s;W)) +
\lambda\|W\|_p$.
$\lambda$ is the constant that controls the capacity of $W$. 
To find the optimum, we follow the previous work and use an optimization scheme
 based on Forward-Backward Splitting (FOBOS)~\cite{fobos}. We experiment with two
 regularization schemes, $p=2$ or the euclidean norm and $p=_*$  or the
 trace norm. In our experiments we found that both the norms have approximately
 similar performance, however the trace norm regularized $W$ has lower capacity and
 hence, is less number of parameters. This is also observed
 by~\cite{bach2008consistency,madhyastha14,madhyastha2014tailoring}.

A by-product of regularizing with trace norm is that we obtain low-dimensional,
aligned-compressed embeddings for both languages. This is possible because of the induced 
low-dimensional properties of $W$. That is, assume $W$ has rank $k$, where $k<n$, such that 
$W \approx U_kV_k^\top$, then the product
$\phi_{s}(e)^{\top}U_kV_k^{\top}\phi_{t}(f)$ gives us
$\phi_{s}(e)^{\top}U_k$ and
$V_k^{\top}\phi_{t}(f)$ compressed embeddings with shared properties. 
We leave exploration of the compressed embeddings for future work.

% \paragraph{Integrating Probabilistic List into the SMT System.}
% By definition, OOV words do no appear in the parallel training corpus
%See (BLIND CITE) for details about optimization of a related log-linear model.

\section{Experiments}
\label{s:exp}

\paragraph{Data and System Settings.}  
For estimating the word embeddings we use the CBOW
algorithm as implemented in the {\tt Word2Vec}
package~\cite{word2vec}\footnote{\url{https://code.google.com/archive/p/word2vec/}} using a
5-token window. We obtain 300 dimension vectors for English and Spanish from a
Wikipedia dump of 2015%
\footnote{Dumps downloaded in January 2015 from \url{https://dumps.wikimedia.org}.}, 
and the Quest
% data\footnote{\url{http://statmt.org/~buck/wmt13qe/wmt13qe_t13_t2_MT_corpus.tgz}}
data\footnote{\url{http://goo.gl/72LLXN}}
which includes subcorpora such as United Nations and Europarl. The final corpus
contains 2.27 billion tokens for English and 840 million tokens for Spanish.
We obtain a coverage of 97\% of the words in our test sets. We also 
remove any occurrence of sentences from the test set that are contained in our
corpus, and avoid any transduction based knowledge transfer. 

To train the log-bilinear softmax based model, we use the dictionary from
the {\tt Apertium project}\footnote{The bilingual dictionary can be downloaded here:
\url{http://goo.gl/TjH31q}.}~\cite{forcada2011apertium}. The dictionary contains
37651 words, we used 70\% of them for training the log-bilinear model and
30\% as a development set for model selection. The average precision @1 was 85.66\% for the
best model over the dev set.   

On the other hand, we build a state-of-the-art phrase-based SMT system trained on
the standard Europarl corpus  for the  English-to-Spanish language pair. 
% (2 million sentences and 54 million tokens) 
% The monolingual corpora used to estimate the WVM is also used to train an additional and larger 
We use a 5-gram language model that is estimated on the target side of the corpus using
interpolated Kneser-Ney discounting with {\tt SRILM}~\cite{srilm}. Additional
monolingual data available within Quest corpora is used to build a larger
language model with the same characteristics. Word alignment is done with {\tt
GIZA++}~\cite{giza} and both phrase extraction and decoding are done with the
{\tt Moses} package~\cite{moses:07}. 

At decoding time, {\tt Moses} allows to include additional translation pairs with their associated
probabilities to selected words via xml mark-up. 
We take advantage of this feature to add our probabilistic estimations to each OOV. 
Since, by definition, OOV words do no appear in the parallel training corpus, they are not present 
in the translation model either and the new translation options only interact with the language
model.

% The model includes the language model, direct and inverse phrase probabilities, direct and inverse lexical  probabilities, phrase and word penalties, and a non-lexicalised reordering. 
The optimization of the weights of the model with the additional translation options
is trained with MERT~\cite{och:03}
against the BLEU~\cite{papineni:02} evaluation metric on the NewsCommentaries
2012\footnote{\url{http://www.statmt.org/wmt13/translation-task.html}} 
(NewsDev) set. We test our systems on the NewsCommentaries 2013 set (NewsTest)
for an in-domain evaluation and on a test set extracted from Wikipedia by Smith
\emph{et.\,al.} \shortcite{smithEtal:2010} for an out-of-domain evaluation
(WikiTest). 

The \emph{domainess} of the test set is established with respect to the number of OOVs. 
Table~\ref{tab:sets} shows the figures of these sets paying special attention
to the OOVs in the basic SMT system. Less than a 3\% of the tokens are OOVs for News data 
(OOV$_{\rm all}$), whereas it is more than a 7\% for Wikipedia's. In our experiments, 
we distinguish between OOVs that are named entities and the rest of content words 
(OOV$_{\rm CW}$). Only about 0.5\% (NewsTest) and 1.8\% (WikiTest) of the tokens fall 
into this category, but we show that they are relevant for the final performance.

\begin{table}[t]
\caption{OOVs on the dev and test sets.}
\label{tab:sets}
\begin{center}
\small
\vspace{-1em}
\begin{tabular}{l rrrr}
\toprule
%  & \mc{4}{c}{English-to-Spanish} & \mc{4}{c}{English-to-German}\\
%  & \mc{1}{c}{sent.} & \mc{1}{c}{tokens} & \mc{1}{c}{OOV$_{\rm all}$} & \mc{1}{l}{OOV$_{\rm CW}$}  
 & \mc{1}{c}{Sent.} & \mc{1}{c}{Tokens} & \mc{1}{c}{OOV$_{\rm all}$} & \mc{1}{c}{OOV$_{\rm CW}$} \\
\midrule
NewsDev  & 3003 & 72988 & 1920 (2.6\%) & 378 (0.5\%) \\
NewsTest & 3000 & 64810 & 1590 (2.5\%) & 296 (0.5\%) \\
WikiTest &  500 & 11069 &  798 (7.2\%) & 201 (1.8\%) \\
\bottomrule
\end{tabular}
\end{center}
\end{table} 
% \vspace*{-\baselineskip}

\paragraph{Evaluation.}
We consider two baseline systems,
the first one does not output any translation for OOVs (\emph{noOOV}), it just ignores the token; 
the second one outputs a verbatim copy of the unseen word as a translation
(\emph{verbatimOOV}). Table~\ref{tab:eval} shows the performance of these systems under three
widely used evaluation metrics TER~\cite{TER}, BLEU and METEOR-paraphrase (MTR)~\cite{METEOR}. 
Including the verbatim copy improves all the lexical evaluation metrics. Specially for named entities 
and acronyms (the 80\% of OOVs in our sets), this is a hard baseline to beat since in most cases 
the same word is the correct translation (e.g. Messi, PHP, Sputnik...).

Next, we enrich the systemns with information gathered from the large monolingual corpora in two
ways, using a bigger language model (\emph{BLM}) and using our newly proposed 
log-bilinear model that uses word embeddings (\emph{BWE}). 
BLMs are very important to improve the fluency of the translations, however they may not be 
helpful for resolving out-of-vocabulary words. 
On the other hand, BWEs are important to make available to the decoder new vocabulary on the topic 
of the otherwise OOVs. Given the large percentage of named entities in the test sets 
(Table~\ref{tab:sets}), our models add the source word as an additional option to the list of 
target words to mimic the \emph{verbatimOOV} system.

\begin{table}[t]
\caption{Automatic evaluation of the translation systems defined in Section~\ref{tab:eval}. The best system is bold-faced (see text for statistical significance).}
\label{tab:eval}
\begin{center}
\small
\vspace{-1em}
\resizebox{\linewidth}{!}{
\begin{tabular}{l c@{\hspace{0.4em}}c@{\hspace{0.4em}}c c@{\hspace{0.4em}}c@{\hspace{0.4em}}c}
 \toprule
 & \mc{3}{c}{NewsTest} & \mc{3}{c}{WikiTest}\\
 & \mc{1}{c@{\hspace{0em}}}{TER} & \mc{1}{c@{\hspace{0em}}}{\hspace{-0.5em}BLEU} & \mc{1}{c@{\hspace{0em}}}{\hspace{-1em}MTR} 
 & \mc{1}{c@{\hspace{0em}}}{TER} & \mc{1}{c@{\hspace{0em}}}{\hspace{-0.5em}BLEU} & \mc{1}{c@{\hspace{0em}}}{\hspace{-1em}MTR} \\
 \midrule
 noOOV 			 & 58.21 & 21.94 & 45.79               & 61.26 & 16.24 & 38.76  \\ 
 verbatimOOV 		 & 57.90 & 22.89 & 47.06               & 58.55 & 21.90 & 45.77  \\ 
 \midrule
 BWE	 	 	 & 58.33 & 22.23 & 45.76                & 58.38 & 21.96 & 44.84  \\ 
 BWE$_{\rm CW50}$	 & 57.66 & 23.09 & 47.14                & 56.19 & 24.16 & 48.49  \\ 
 BWE$_{\rm CW10}$	 & 57.85 & 23.06 & 47.11                & 55.64 & 24.71 & 49.05  \\ 
 BLM	 		 & 55.37 & 25.83 & {\bf 49.19}          & 52.60 & 30.63 & 51.04  \\ 
 BLM+BWE	       	 & 55.89 & 24.92 & 47.84                & 51.02 & 32.20 & 52.09  \\ 
 BLM+BWE$_{\rm 50}$  	 & 55.55 & 25.61 & 49.01                & 49.50 & 33.94 & 54.93  \\ 
 BLM+BWE$_{\rm 10}$  	 & {\bf 55.31} & {\bf 25.86} & 49.04    & {\bf 49.12} & {\bf 34.58} & {\bf 55.52}  \\ 
 \bottomrule
\end{tabular}
}
\end{center}
\end{table} 
Table~\ref{tab:eval} includes seven systems with the additional monolingual
information. Three of them add, at decoding time, the top-$n$ translation
options given by the BWE for a OOV. \emph{BWE} system uses the top-50 for all
the OOVs,  \emph{BWE$_{\rm CW50}$} also uses the top-50 but only for content
words other than named entities\footnote{We consider a named entity any word
that begins with a capital letter and is not after a punctuation mark, and any
fully capitalized word.}, and \emph{BWE$_{\rm CW10}$} limits the list to 10
elements. \emph{BLM}  is the same as the baseline system \emph{verbatimOOV} but
with the large language model. \emph{BLM+BWE}, \emph{BLM+BWE$_{\rm 50}$} and
\emph{BLM+BWE$_{\rm 10}$} combine the three BWE systems with the large language
model.

A large number of unseen words in the NewsTest are mostly named entities, using BWEs
to translate all the words, including named entities, barely improves the translation. Also, the richness in
vocabulary, consisting of many names, adds noise to the decoder. 
We observe that the improvements are moderate in the NewsTest (in-domain
dataset), this mostly is  
because the differences in the probability of the BWE translation options are
very small owing to the candidates
being named entities. We also see that this affects the overall
integration of the scores into the decoder and also induces ambiguity in
the system. 
On the other hand, we observe that the decoder benefits from the
information on content words, specially for the out-of-domain WikiTest set,
given the constrained list of alternative translations (\emph{BWE$_{\rm CW10}$}
achieves 2.75 BLEU points of improvement).

The addition of the large language model improves the results significantly.
When combined with the BWEs we observe that the  BWEs clearly help in the
translation of WikiTest but do not seem as relevant in the
in-domain set. We also achieve a statistically significant improvement of 3.9 points of BLEU with
the BLM and BWE combo
system in WikiTest ($p<0.001$). The
number of translation options in the list is also relevant, we see that for
\emph{BWE$_{\rm CW50}$} we have an improvement of 3.3 points on BLEU. 
We also observe that the results
are consistent among different metrics.

We have further manually evaluated the translation of WikiTest using \emph{BWE$_{\rm
CW50}$}. We obtained an accuracy of a 68\%, that is, the BWE
gives the correct translation option at least 68\% of the times. The other 32\%
of time, it fails as the words in the translated language happened to be either
multiwords or named entities.
In table~\ref{tab:examples} we observe some of the
these examples. The first two examples \emph{galaxy} and \emph{nymphs}
are nouns where we obtain the first option as the correct translation. The problem is
harder for named entities as we observe in the table,
the name \emph{Stuart} in English has \emph{William} as
most probable translation in Spanish, the correct translation
\emph{Estuardo}
however appears as the 48th choice. Our model is also unable to generate
multiword expressions, as shown in the table for the english word
\emph{folksong}, the correct translation being \emph{canci\'on folk}. This would need two words in Spanish in order to be translated properly, 
however, our model does obtain words: \emph{canci\'on} and \emph{folclore} as
the most probable translation options.

% \vspace*{-\baselineskip}
\begin{table}[t]
\caption{Top-$n$ list of translations obtained with the bilingual embeddings.}
\label{tab:examples}
\begin{center}
 \small
 \vspace{-1em}
\begin{tabular}{l@{\hspace{1em}}l@{\hspace{1em}}l@{\hspace{1em}}l}
\toprule
\sc{galaxy}   & \sc{nymphs} & \sc{Stuart} & \sc{folksong}      \\
\midrule
\bf{galaxia}  & \bf{ninfas} & William     & m\'usica           \\
planeta       & ninfa       & Henry       & \bf{folclore}      \\
universo      & cr\'ias     & John        & literatura         \\
planetas      & diosa       & Charles     & himno              \\
galaxias      & dioses      & Thomas      & folklore           \\
%     ... &       ...        & ...         &      ...             \\
     ...   &       ...           & Estuardo (\#48) & \bf{canci\'on} (\#7) \\
% galaxia  (0.0353) & m\'usica (0.0230)   & William (0.0256) \\
% planeta  (0.0318) & folclore (0.0229)   & Henry   (0.0241) \\
% universo (0.0277) & literatura (0.0228) & John    (0.0236) \\
% planetas (0.0266) & himno (0.0227)      & Charles (0.0236) \\
% galaxias (0.0265) & folklore (0.0227)   & Thomas  (0.0234) \\
%     ... & ... & ...  \\
%         & canci\'on (0.0224) & Estuardo (0.0180) \\
\bottomrule
\end{tabular}
\end{center}
\end{table}

\section{Conclusions}
We have presented a method for resolving unseen words in SMT that performs
vocabulary expansion by using a simple log-bilinear softmax based model.
The
addition of translation options to a mere 1.8\% of the words has allowed the
system to obtain a
relative improvement of a 13\% in BLEU (3.9 points) for out-of-domain data. For
in-domain data, where the number of content words is small, improvements
are moderate. 
We would like to further study the repercussion of this simple method on diverse
and most distant language pairs and how the form of the loss function affects
the quality of the bilingual word embeddings. 
%A margin-loss instead of the
%current log-linear function could further help in improving the prediction.

\bibliography{refs}
\bibliographystyle{emnlp2016}

\end{document}